\documentclass{IEEEtran}
\usepackage{cite}
\usepackage{amsmath,amssymb,amsfonts}
\usepackage{graphicx}
\usepackage{textcomp,nicefrac}
\usepackage{caption} 
\def\BibTeX{{\rm B\kern-.05em{\sc i\kern-.025em b}\kern-.08em
T\kern-.1667em\lower.7ex\hbox{E}\kern-.125emX}}
\begin{document}
\title{Few-Shot Classification and Anatomical Localization of Tissues in SPECT Imaging} 


\author{Mohammed Abdul Hafeez Khan, \IEEEmembership{Graduate Student Member, IEEE}, Samuel Morries Boddepalli, \IEEEmembership{Graduate Student Member, IEEE}, Siddhartha Bhattacharyya, \IEEEmembership{Senior Member, IEEE}, and Debasis Mitra, \IEEEmembership{Senior Member, IEEE}
\thanks{Authors Mohammed Abdul Hafeez Khan, Samuel Morries Boddepalli, Siddhartha Bhattacharyya and Debasis Mitra are all affiliated with Florida Institute of Technology, Melbourne, FL 32901 USA (emails:mkhan@my.fit.edu, sboddepalli2023@my.fit.edu, sbhattacharyya@fit.edu and dmitra@fit.edu)}}

\maketitle

\begin{abstract}
Accurate classification and anatomical localization are essential for effective medical diagnostics and research, which may be efficiently performed using deep learning techniques. However, availability of limited labeled data poses a significant challenge. To address this, we adapted Prototypical Networks and the Propagation-Reconstruction Network (PRNet) for few-shot classification and localization,  respectively, in Single Photon
Emission Computed Tomography (SPECT) images. For the proof of concept we used a 2D-sliced image cropped around heart. The Prototypical Network, with a pre-trained ResNet-18 backbone, classified ventricles, myocardium, and liver tissues with 96.67\% training and 93.33\% validation accuracy. PRNet, adapted for 2D imaging with an encoder-decoder architecture and skip connections, achieved a training loss of 1.395, accurately reconstructing patches and capturing spatial relationships. These results highlight the potential of Prototypical Networks for tissue classification with limited labeled data and PRNet for anatomical landmark localization, paving the way for improved performance in deep learning frameworks.
\end{abstract}


\section{Introduction}
\label{sec:introduction}

\IEEEPARstart{I}{n} the rapidly advancing domain of medical imaging, accurate classification and anatomical localization of tissues are critical for diagnostic, therapeutic, and research purposes. Recent advances in the application of deep learning techniques have significantly expanded their potential in imaging classification \cite{ker2017deep}\cite{khan2022detection}. However, the scarcity of data in the medical domain often necessitates specialized adaptations to meet the stringent demands for precision and reproducibility. This research draws inspiration from two pivotal studies: The development of Prototypical Networks \cite{snell2017prototypical}, and Propagation Reconstruction Network (PRNet) \cite{lei2023one}. These studies form the foundation for our methodologies, adapted and applied to Single Photon Emission Computed Tomography (SPECT) images, focusing on the tissues of the ventricles, myocardium, and liver.

\begin{figure}[b]
    \centering
    \includegraphics[width=0.6\columnwidth]{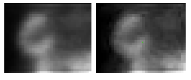} 
    \caption{Real vs reconstructed image of myocardium tissue using PRNet}
    \label{fig:prnet}
\end{figure}

The first aim of our study was to address the challenges posed by a severely limited dataset of only 12 SPECT images. To address this shortcoming, we implemented a Few-Shot Learning (FSL) approach, leveraging Prototypical Networks \cite{snell2017prototypical}. This approach was tailored to classify segmented tissue mask types—myocardium, ventricles, and liver—by representing each class with a prototype in a learned metric space. The second aim of our study was to accurately determine the anatomical locations of tissues within 2D SPECT images. To achieve this, we adapted the architecture of the PRNet \cite{lei2023one}, originally designed for 3D CT and MRI images, for 2D convolutional processing. As shown in Fig. \ref{fig:prnet}, we accurately reconstruct patches, capturing spatial relationships to predict the relative spatial coordinates between anatomical landmarks. We reviewed advancements in few-shot learning that have shaped image classification. Koch et al. \cite{koch2015siamese} used Siamese Neural Networks for one-shot image recognition, setting a benchmark for learning with minimal data. Building on this, Shaban et al. \cite{shaban2017one} enhanced semantic segmentation by adapting Fully Convolutional Networks (FCNs) to new tasks. Similarly, Snell et al. \cite{snell2017prototypical} introduced Prototypical Networks to streamline classification by creating class-specific prototypes in a learned metric space. Expanding on these ideas, Rakelly et al. \cite{rakelly2018conditional} used support images to improve image classification.


\section{Methodology}
\label{sec:methodology}

\subsection{Prototypical Network}
To apply Prototypical Network for few-shot classification of myocardium, ventricles, and liver masks, we designed a convolutional neural network (CNN) to map input images to a metric space via the embedding function \( f_\phi \). We used a pretrained ResNet-18\cite{chen2019closer} as the backbone, fine-tuning it for the specific few-shot classification task. The forward pass of the Prototypical Network computes classification scores using support and query images. The support images create prototype \( c_k \) representations for each class \(k\), which are the mean feature vectors of the support examples. Query images \( x \) are then classified by computing a softmax over the Euclidean distances \(d(z,z')\) to these prototypes, as shown in Eq. \ref{eq5}:
\begin{equation}
\label{eq5}
p_\phi(y = k | x) = \frac{\exp(-d(f_\phi(x), c_k))}{\sum_{k'} \exp(-d(f_\phi(x), c_{k'}))} 
\end{equation}
Subsequently, using the Eq. \ref{eq6}, the classification scores are obtained by minimizing the negative log-probability.  
\begin{equation}
\label{eq6}
J(\phi) = - \log p_\phi(y = k | x)
\end{equation}
The model then selects the class \(k\) with the highest score for each query image. Accuracy is calculated by comparing the predicted class with the true label and dividing the number of correct predictions by the total predictions. 

Therefore, prototypical networks map input data to an embedding space and classify using class prototypes, enabling efficient classification from just a few samples of data.

\subsection{Propagation-Reconstruction Network (PRNet)}

The Propagation-Reconstruction Network (PRNet) was originally designed for 3D CT and MRI images \cite{lei2023one}. We adapted it for 2D slices of SPECT imaging by modifying its configuration from 3D to 2D layers, enabling it to focus on spatial information and accurately identify anatomical landmarks. PRNet includes an encoder, fully connected layers for predicting anatomical positions, and a decoder that reconstructs the original image for validation, as shown in Fig. \ref{fig:prnet}. The model uses self-supervised learning by randomly selecting two points, \(c_i\) and \(c_j\), within an image. The relative offset between them, \(d_{ji}\), guides the learning process. Two fixed-size patches, \(x(c_i)\) and \(x(c_j)\), are cropped around \(c_i\) and \(c_j\), then processed through PRNet to obtain their 2D anatomical coordinates, \(a(c_i)\) and \(a(c_j)\). The predicted offset, \(d'_{ji}\), determines the spatial relationship between the points in the latent vector space. As shown in Eq. \ref{eq1}, PRNet employs a self-supervised loss function \(L_{ssl}\), composed of two main components: the distance loss (\( L_{dis} \)), which quantifies the error in the predicted relative positions, and the reconstruction loss (\( L_{rec} \)), which measures the accuracy of the reconstructed patches x\(_r(c_i)\) and x\(_r(c_j)\), compared to the originally cropped patches.

\begin{equation}
\label{eq1}
L_{ssl} = L_{dis} + L_{rec}
\end{equation}

In summary, PRNet provides a self-supervised framework that leverages inherent anatomical similarities across individuals to facilitate landmark localization in medical imaging with minimal reliance on extensive labeled datasets.


\section{Results and Discussion}
\label{sec:expresults}

The experimental results highlight the effectiveness of both the Prototypical Network and PRNet in their respective tasks.

\begin{figure}[t]
    \centering
    \includegraphics[width=0.7\columnwidth]{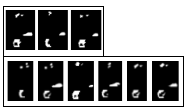} 
    \caption{Support (N=3) and query (N=6) sets of segmented tissues for training Prototypical Network, which needs labeled support set and unlabeled query set}
    \label{fig:proto}
\end{figure}

For the Prototypical Network, we used episodic batches for training and testing. Each episode included a fixed number of classes, support images, and query images. As illustrated in Fig. \ref{fig:proto}, we used \textbf{3} support images and \textbf{6} query images per class, training the model for 10 episodes with the Adam optimizer (learning rate: \textbf{0.001} and cross-entropy loss). The model achieved \textbf{96.67\%} accuracy and \textbf{0.486} loss on the training set, and \textbf{93.33\%} accuracy and \textbf{0.366} loss on the validation set. It effectively learned to classify tissue types from limited labeled data, demonstrating its proficiency in few-shot learning for classification of SPECT images.

The PRNet model, trained using numerous pairs of cropped patches from a single SPECT image via random double cropping, used an encoder-decoder architecture with skip connections to learn relative positions between patches. Training involved minimizing the self-supervised loss \( L_{ssl} \) using the Adam optimizer (learning rate: \textbf{1e-3)}. Over \textbf{50} epochs, each with 10 iterations, the model achieved a training loss of \textbf{1.395}, demonstrating its ability to capture spatial relationships and accurately reconstruct input patches, as illustrated in Fig. \ref{fig:prnet}.

\section{Conclusion}
\label{sec:conclusion}

In this paper, we addressed the challenges of accurate tissue classification and anatomical localization in SPECT imaging under the constraint of limited data. The Prototypical Network effectively classified myocardium, ventricle, and liver tissues with high training and validation accuracies, demonstrating proficiency in few-shot learning. PRNet, adapted for SPECT imaging, learned spatial relationships between anatomical landmarks and achieved a low training loss, paving the way for integrating and improving segmentation accuracy in deep learning frameworks in the future.

{\em Acknowledgement:} This work was performed as a class project in the Artificial Intelligence course at the Florida Institute of Technology in Spring 2024 and was partially supported by the NIH grant R15EB030807.

\bibliographystyle{IEEEtran}
\bibliography{references.bib}

\begin{thebibliography}{1}
\providecommand{\url}[1]{#1}
\csname url@samestyle\endcsname
\providecommand{\newblock}{\relax}
\providecommand{\bibinfo}[2]{#2}
\providecommand{\BIBentrySTDinterwordspacing}{\spaceskip=0pt\relax}
\providecommand{\BIBentryALTinterwordstretchfactor}{4}
\providecommand{\BIBentryALTinterwordspacing}{\spaceskip=\fontdimen2\font plus
\BIBentryALTinterwordstretchfactor\fontdimen3\font minus \fontdimen4\font\relax}
\providecommand{\BIBforeignlanguage}[2]{{%
\expandafter\ifx\csname l@#1\endcsname\relax
\typeout{** WARNING: IEEEtran.bst: No hyphenation pattern has been}%
\typeout{** loaded for the language `#1'. Using the pattern for}%
\typeout{** the default language instead.}%
\else
\language=\csname l@#1\endcsname
\fi
#2}}
\providecommand{\BIBdecl}{\relax}
\BIBdecl

\bibitem{ker2017deep}
J.~Ker, L.~Wang, J.~Rao, and T.~Lim, ``Deep learning applications in medical image analysis,'' \emph{Ieee Access}, vol.~6, pp. 9375--9389, 2017.

\bibitem{khan2022detection}
M.~H. Khan, P.~S. Giri, and J.~A.~A. Jothi, ``Detection of cavities from oral images using convolutional neural networks,'' in \emph{2022 International Conference on Electrical, Computer and Energy Technologies (ICECET)}.\hskip 1em plus 0.5em minus 0.4em\relax IEEE, 2022, pp. 1--6.

\bibitem{snell2017prototypical}
J.~Snell, K.~Swersky, and R.~Zemel, ``Prototypical networks for few-shot learning,'' \emph{Advances in neural information processing systems}, vol.~30, 2017.

\bibitem{lei2023one}
W.~Lei, Q.~Su, T.~Jiang, R.~Gu, N.~Wang, X.~Liu, G.~Wang, X.~Zhang, and S.~Zhang, ``One-shot weakly-supervised segmentation in 3d medical images,'' \emph{IEEE Transactions on Medical Imaging}, 2023.

\bibitem{koch2015siamese}
G.~Koch, R.~Zemel, R.~Salakhutdinov \emph{et~al.}, ``Siamese neural networks for one-shot image recognition,'' in \emph{ICML deep learning workshop}, vol.~2, no.~1.\hskip 1em plus 0.5em minus 0.4em\relax Lille, 2015.

\bibitem{shaban2017one}
A.~Shaban, S.~Bansal, Z.~Liu, I.~Essa, and B.~Boots, ``One-shot learning for semantic segmentation,'' \emph{arXiv preprint arXiv:1709.03410}, 2017.

\bibitem{rakelly2018conditional}
K.~Rakelly, E.~Shelhamer, T.~Darrell, A.~Efros, and S.~Levine, ``Conditional networks for few-shot semantic segmentation,'' 2018.

\bibitem{chen2019closer}
W.-Y. Chen, Y.-C. Liu, Z.~Kira, Y.-C.~F. Wang, and J.-B. Huang, ``A closer look at few-shot classification,'' \emph{arXiv preprint arXiv:1904.04232}, 2019.

\end{thebibliography}

\end{document}